# Shift-Invariant Sparse Coding for Audio Classification

**Roger Grosse**
Computer Science Dept.
Stanford University
Stanford, CA 94305

**Rajat Raina**
Computer Science Dept.
Stanford University
Stanford, CA 94305

**Helen Kwong**
Symbolic Systems Program
Stanford University
Stanford, CA 94305

**Andrew Y. Ng**
Computer Science Dept.
Stanford University
Stanford, CA 94305

## Abstract

Sparse coding is an unsupervised learning algorithm that learns a succinct high-level representation of the inputs given only unlabeled data; it represents each input as a sparse linear combination of a set of basis functions. Originally applied to modeling the human visual cortex, sparse coding has also been shown to be useful for self-taught learning, in which the goal is to solve a supervised classification task given access to additional unlabeled data drawn from *different* classes than that in the supervised learning problem. Shift-invariant sparse coding (SISC) is an extension of sparse coding which reconstructs a (usually time-series) input using all of the basis functions in all possible shifts. In this paper, we present an efficient algorithm for learning SISC bases. Our method is based on iteratively solving two large convex optimization problems: The first, which computes the linear coefficients, is an $L_1$-regularized linear least squares problem with potentially hundreds of thousands of variables. Existing methods typically use a heuristic to select a small subset of the variables to optimize, but we present a way to efficiently compute the exact solution. The second, which solves for bases, is a constrained linear least squares problem. By optimizing over complex-valued variables in the Fourier domain, we reduce the coupling between the different variables, allowing the problem to be solved efficiently. We show that SISC's learned high-level representations of speech and music provide useful features for classification tasks within those domains. When applied to classification, under certain conditions the learned features outperform state of the art spectral and cepstral features.

## 1 Introduction

In supervised learning, labeled data is often difficult and expensive to obtain. Thus, several recent machine learning models attempt to use additional sources of data to improve performance. To motivate later discussion, consider a speaker identification task in which we'd like to distinguish between Adam and Bob from their speech, given only a small amount of labeled training data. Semi-supervised learning [1] algorithms can make use of additional unlabeled data that has the same class labels as the classification task. Thus, to apply most semi-supervised learning methods (such as [2]) to distinguish between Adam and Bob, we must obtain additional unlabeled data from Adam and Bob specifically. For many problems, this sort of unlabeled data is hard to obtain; indeed, short of eavesdropping on conversations between Adam and Bob, it is hard to envisage a data collection method that results in unlabeled speech samples from Adam and Bob—*and no one else*—that does not also automatically result in our having the class labels as well. Transfer learning [3, 4, 5] typically attempts to use additional labeled data to construct supervised classification problems that are related to the task of interest. Thus, for speaker identification, transfer learning may require labeled data from other speech classification tasks. Since these additional classification tasks are supervised and thus require labeled examples, data for even transfer learning is often difficult to obtain.[1]

In contrast, the recently introduced self-taught learning framework attempts to use easily obtainable unlabeled data to improve classification performance. In particular, in self-taught learning the unlabeled data does not have to share the labels of the classification task, and in general does not even have to arise from the latter's generative distribution. [6, 7] For the task of distinguishing between Adam and Bob, this means we may use unlabeled data from speakers *other* than Adam and Bob (or perhaps even use non-speech audio samples). Given the ease with which such data can be obtained—one can envisage recording audio from radio broadcasts or downloading audio clips randomly off the internet—we believe that self-taught learning holds the promise of much easier applicability than most semi-supervised learning and transfer learning algorithms.

---

[1] In some cases, the additional classification tasks can be automatically constructed using ingenious hand-engineered heuristics [5].

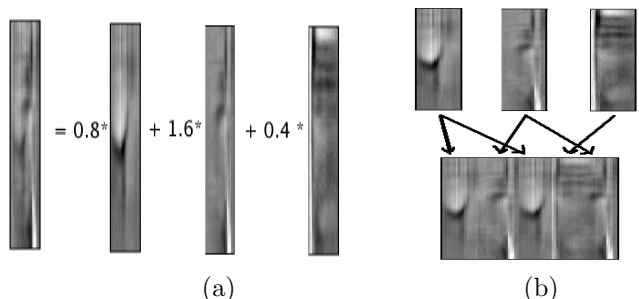

(a) (b)

Figure 1: Cartoon illustrations of sparse coding. (a) Regular sparse coding applied to the spectrogram of an input signal. The input (left) is represented approximately as a linear combination of a small number of basis functions (spectrograms). (b) Shift-invariant sparse coding allows the input (bottom) to be represented by reusing the same basis functions (top) at different time offsets in the signal. Each arrow represents a nonzero coefficient corresponding to a basis and a time offset.

The issue of what data there is to learn from lies at the heart of all machine learning methods. In supervised learning, even an inferior learning algorithm can often outperform a superior one if it is given more data. Self-taught learning uses a type of unlabeled data (that does not share the labels of the classification task of interest) that is often easily obtained even in massive quantities, and that thus can provide a large number of "bits" of information for algorithms to try to learn from. Thus, we believe that if good self-taught learning algorithms can be developed, they hold the potential to make machine learning significantly more effective for many problems. In this paper, we present an algorithm for self-taught learning based on shift-invariant sparse coding (SISC), and apply it to audio classification.

## 2 Overview

Our algorithm for self-taught learning of audio signals is based on the principle of sparse coding (first introduced by Olshausen & Field [8]), which attempts to discover a representation of the input signals that is *sparse*—i.e., a representation in which most of the components are zero. Specifically, our algorithm first attempts to learn a (possibly large) dictionary of patterns (audio snippets) called "basis functions," such that the input audio signals can be represented approximately as a superposition of only a small number of "active" basis functions. For any given input signal, these active basis functions produce a sparse signal representation that concisely represents that signal.

For illustration, consider a 500ms speech signal sampled at 16kHz. The raw acoustic signal is a vector $x \in \mathbb{R}^{8000}$. This unwieldy representation can be extremely difficult to reason over. Instead, our self-taught learning algorithm discovers statistical correlations from unrelated unlabeled data, and thus learns a dictionary of 200 basis functions (say) that, when superposed only a few at a time, suffice to adequately describe most input audio signals. Given a 500ms speech signal, the algorithm might be able to represent the signal using just 10 (say) of these learnt basis functions. Informally, the algorithm is able to describe the 8000 numbers succinctly using only 10 numbers corresponding to the activated bases. As we demonstrate in our experiments, such a succinct representation can be significantly easier to reason with. (See Figure 1a.)

In the example of speech, we can informally think of the audio speech signal as being generated by a small number of distinct acoustic events, such as vibrations of the vocal cords. These underlying acoustic events might provide a concise and relatively noiseless representation of the signal; they also provide a slightly higher-level representation that might be easier to reason with than the raw acoustic signal.[2]

In this paper, we attempt to automatically discover such a higher-level representation using only unlabeled audio signals. As in [6, 7], our algorithms will first learn a large dictionary of patterns (or basis functions), and then reconstruct any new input signal using a weighted combination of a few of these basis functions. The weights (and positions, or shifts) of these basis functions then give a slightly higher-level, and more succinct, representation of the input; this representation can then be used in supervised learning.

When sparse coding is applied to audio signals, there is a significant technical problem: sparse coding is required to learn the occurrence of each acoustic pattern *at each time offset* as a wholly different basis function. This is clearly undesirable—in the example of speech data, we want a basis to be capable of representing a particular phone at *any* time in the signal. Shift-invariant sparse coding (SISC) is an extension of sparse coding that allows each basis function to be replicated at each time offset within the signal (Figure 1b). [10, 11] However, current algorithms for SISC rely on heuristic solutions, and finding exact

---

[2]For comparison, it is also well-known that visual signals can be encoded efficiently as a combination of only a few oriented edges called Gabor filters, that also resemble the receptive fields of certain cells in the primary visual cortex. [9, 8]

solutions has been intractable. [10, 11, 12] In neuroscience, such heuristically-computed variants of SISC have been used to model the responses of cells in the cochlea (ear) and auditory nerve [11]; for music, they have also been used to separate musical instruments in an audio recording. [12] We present an efficient algorithm for computing SISC solutions, and apply it to self-taught learning.

## 2.1 Preliminaries and notation

To motivate our definition of the shift-invariant sparse coding (SISC) problem, we first introduce the standard sparse coding model. Sparse coding can be expressed as a generative model in which the $m$ unlabeled input signals $x^{(i)} \in \mathbb{R}^p, i = 1, \ldots, m$ are assumed to be generated as linear combinations of some dictionary of $n$ basis functions, $a^{(j)} \in \mathbb{R}^p, j = 1, \ldots, n$, with additive Gaussian noise $\epsilon^{(i)} \in \mathbb{R}^p$:

$$\begin{aligned} x^{(i)} &= \sum_{j=1}^{n} s^{(i,j)} a^{(j)} + \epsilon^{(i)} \\ s^{(i,j)} &\sim Laplacian(1/\beta) \\ \epsilon^{(i)} &\sim \mathcal{N}(0, \sigma^2 I) \end{aligned}$$

where $s^{(i,j)} \in \mathbb{R}$ are the coefficients of the linear combination, and are drawn from a Laplacian (or other heavy-tailed) prior distribution with parameter $\beta$. For example, when each input is a 500ms speech signal sampled at 16kHz, each input $x^{(i)} \in \mathbb{R}^{8000}$ and basis vector $a^{(j)} \in \mathbb{R}^{8000}$ represents time-domain amplitudes. Given only unlabeled inputs $x^{(i)}$, sparse coding poses the following optimization problem to compute the maximum-a-posteriori (MAP) estimates of both the bases $a^{(j)}$ and coefficients $s^{(i,j)}$:[3]

$$\begin{aligned} \min_{a,s} \quad & \sum_{i=1}^{m} \|x^{(i)} - \sum_{j=1}^{n} a^{(j)} s^{(i,j)}\|_2^2 \\ & + \beta \sum_{i,j} |s^{(i,j)}| \quad (1) \\ \text{s.t.} \quad & \|a^{(j)}\|_2^2 \leq c, \quad 1 \leq j \leq n. \quad (2) \end{aligned}$$

The normalization constraint in (2) prevents trivial solutions where $s^{(i,j)}$ becomes very small and $a^{(j)}$ becomes very large. This problem is convex in either $s$ or $a$ (though not jointly convex in both). So, we can alternately solve two convex optimization problems: an $L_1$-regularized least squares problem solving for the coefficients $s$ keeping the bases $a$ fixed, and an $L_2$-constrained convex problem solving for the bases $a$ keeping the coefficients $s$ fixed. [13]

In shift-invariant sparse coding (SISC), each basis function is allowed to appear at all possible shifts within the signal. [10, 11] To use ideas recently developed for solving the sparse coding model above, we pose the SISC problem as the natural extension to the sparse coding optimization problem (1-2):

$$\begin{aligned} x^{(i)} &= \sum_{j} a^{(j)} * s^{(i,j)} + \epsilon^{(i)} \\ s^{(i,j)} &\sim Laplacian(1/\beta) \\ \epsilon^{(i)} &\sim \mathcal{N}(0, \sigma^2 I) \end{aligned}$$

where we use the convolution operator $*$ to succinctly represent each basis $a^{(j)}$ being used at all possible time shifts within the signal $x^{(i)}$.[4] Here, $x^{(i)} \in \mathbb{R}^p$ as before, but we allow the bases $a^{(j)} \in \mathbb{R}^q$ to be of a lower dimension than the input signal itself ($q \leq p$). In our earlier example, if each input $x^{(i)} \in \mathbb{R}^{8000}$ is a 500ms speech signal, we might choose to avoid redundancy by using 100ms basis functions, in which case $a^{(j)} \in \mathbb{R}^{1600}$. Further, the coefficient $s^{(i,j)}$ corresponding to the input $x^{(i)}$ and basis $a^{(j)}$ is now a vector $s^{(i,j)} \in \mathbb{R}^{p-q+1}$, representing the coefficient for each possible temporal offset of the basis $a^{(j)}$. We will use the notation $s_t^{(i,j)}$ to represent the coefficient corresponding to the input $x^{(i)}$, basis $a^{(j)}$ and temporal offset $t$.

The corresponding SISC optimization problem for learning bases $a$ and coefficients $s$ then becomes:

$$\begin{aligned} \min_{a,s} \quad & \sum_{i=1}^{m} \|x^{(i)} - \sum_{j=1}^{n} a^{(j)} * s^{(i,j)}\|_2^2 \\ & + \beta \sum_{i,j} \|s^{(i,j)}\|_1 \quad (3) \\ \text{s.t.} \quad & \|a^{(j)}\|_2^2 \leq c, \quad 1 \leq j \leq n \quad (4) \end{aligned}$$

Note that by explicitly representing each coefficient $s_t^{(i,j)}$ and expanding out the convolution, we can, in principle, formulate the above optimization problem as a very large sparse coding problem (1-2) with tied parameters. However, such a reformulation would ignore the special structure in the problem (3-4), and is computationally infeasible to solve for even moderate problem sizes. In this paper, we will present an efficient algorithm for SISC.

## 3 Efficient SISC Algorithm

We solve the optimization problem (3-4) by alternately solving two large convex optimization problems: com-puting the coefficients $s^{(i,j)}$ given a fixed basis set, and finding the bases $a^{(j)}$ given the coefficients.[5] We now describe efficient algorithms for each of these problems.

---

[3]For simplicity, we have set the scale factor $\sigma^2 = 1$, and assumed a uniform prior over the bases $a$.

[4]If $h = f * g$ is the 1-D discrete convolution of two vectors $f \in \mathbb{R}^p$ and $g \in \mathbb{R}^q$, then $h_t = \sum_{\tau=\max(1,t-q+1)}^{\min(p,t)} f_\tau g_{t-\tau+1}$. The dimension of the vector $h$ is given by $\dim(h) = \dim(f) + \dim(g) - 1$. For ease of exposition, we present our SISC model and algorithms only for audio or other sequential (1-D) signals. However, the model and algorithm can also be straightforwardly extended to other input signals with spatial and/or temporal extent (such as images or videos).

## 3.1 Solving for coefficients

Given a fixed set of bases $a$, the optimization problem (3) over coefficients $s$ decomposes over $i = 1, \ldots, m$ into a set of $m$ independent optimization problems. In other words, the coefficients $\{s^{(i,j)}, j = 1, \ldots, n\}$ corresponding to an input $x^{(i)}$ can be found independently of the coefficients corresponding to other inputs. Thus, we will focus on the problem of finding the coefficients $s$ corresponding to a single input $x \in \mathbb{R}^p$:

$$\min_s \|x - \sum_{j=1}^n a^{(j)} * s^{(j)}\|_2^2 + \beta \sum_{j=1}^n \|s^{(j)}\|_1 \quad (5)$$

Here, $s^{(j)} \in \mathbb{R}^{p-q+1}$ represents the coefficients associated with basis $a^{(j)}$. Optimizing (5) with respect to $s^{(j)}$ is difficult because the variables are highly coupled. (For instance, recall that the coefficient $s_t^{(j)}$ for a basis function at one time instant and the coefficient $s_{t+1}^{(j)}$ for the same basis function at the next instant are two distinct variables. A basis function typically correlates quite highly with the same basis function shifted by one timestep, and this makes it hard to determine which of the two coefficients better "explains" the input signal $x$.) Previous work [11, 12] computed coefficients using gradient descent on the objective function (5). Gradient descent does not typically work well with problems whose variables are tightly coupled. Most existing implementations, therefore, use a heuristic to select a subset of the coefficients to optimize, and set the other coefficients to zero. As we demonstrate in the experiments section, the tight coupling between coefficients makes it difficult to heuristically select a small subset of variables to optimize, and as a result, such heuristic techniques typically lead to suboptimal solutions.

Our approach will efficiently find the exact optimum with respect to all the coefficients by taking advantage of the "feature-sign" trick: the $L_1$-regularized least squares problem (5) reduces to an unconstrained quadratic optimization problem over just the nonzero coefficients, if we are able to correctly guess the signs (positive, zero or negative) of the optimal values of the coefficients. Such a quadratic optimization problem can then be easily solved using standard numerical methods. This feature-sign trick has been formalized in the *feature-sign search* algorithm, [13] which greedily searches the space of nonzero coefficients and their signs, and provably converges to the optimal solution.

The feature-sign search algorithm can efficiently find coefficients for short signals (i.e., when $x$ is low-dimensional). For example, if fewer than 1000 coefficients are nonzero over the course of the optimization for problem (5) (and at its optimal solution), then each of the quadratic optimization problems solved in the inner loop of feature-sign search will also involve at most 1000 variables, so that they can be solved efficiently. However, we are often interested in solving for the coefficients corresponding to fairly long signals (such as if $x$ is a 1 minute long speech signal). Empirically, the number of nonzero coefficients in the optimal solution grows roughly linearly with the length of the signal (i.e., the dimensionality of $x$). Since solving each least squares problem takes about $O(Z^3)$ time for $Z$ nonzero coefficients, the standard feature-sign search algorithm becomes expensive for long signals.

---

**Algorithm 1** FS-EXACT

**input** Input $x \in \mathbb{R}^p$, bases $a^{(1)}, \ldots, a^{(n)} \in \mathbb{R}^q$.
**output** Optimal coefficients $s^{(j)} \in \mathbb{R}^{p-q+1}$ for (5).
**algorithm**
    Initialize $s^{(j)} := 0$, or to value from previous iteration.
    **while** true **do**
        $S := \{s_i^{(j)} \neq 0\}$
        $S := S \cup \{\text{up to } k \text{ coefficients not in } S \text{ with the largest (magnitude) partial derivatives of the quadratic term}\}$.
        Solve (5) exactly for $S$ using feature sign search.
        Break if all partial derivatives are smaller than $\beta$.
    **end while**
    **return** the optimal coefficients $s^{(j)}$.

---

**Algorithm 2** FS-WINDOW

**input** Input $x \in \mathbb{R}^p$, bases $a^{(1)}, \ldots, a^{(n)} \in \mathbb{R}^q$. (Assume for simplicity that $p - q + 1$ is divisible by $q$.)
**output** Estimated coefficients $s^{(j)} \in \mathbb{R}^{p-q+1}$ for (5).
**algorithm**
    Initialize $s^{(j)}$ arbitrarily, or to value from previous iteration; $numwindows = (p - 2q + 1)/q$.
    **for** $pass = 1$ to $numpasses$ **do**
        **for** $w = 1$ to $numwindows$ **do**
            $W := \{s_i^{(j)} | q(w-1) + 1 \leq i \leq q(w+1)\}$
            Solve (5) exactly for $W$ using FS-EXACT
        **end for**
    **end for**
    **return** the coefficients $s^{(j)}$.

---

[5] Rather than use entire sentences or songs for the input vectors $x^{(i)}$, we found it was much more efficient to divide the signals into medium-sized excerpts of 500 ms, which would be treated as independent. This way, given a fixed set of bases, the optimal coefficients could be computed efficiently for each excerpt. To reduce edge effects, each segment was multiplied by a window function which decays smoothly to zero. We randomly divided the training data into 2 batches and alternately optimized with respect to each batch.

To address this problem, we use an iterative sliding window approach as shown in Algorithm 2 to solve a series of smaller problems using the feature-sign search algorithm. These smaller problems are generated by attempting to solve for only $2q$ coefficients at one time, while keeping the other coefficients fixed, and iterating till convergence. Since the objective function in problem (5) is convex, this method is guaranteed to converge to the optimal solution. We will refer to plain feature sign (without the sliding window) as FS-EXACT and feature sign with sliding window as FS-WINDOW.

Importantly, even though in the worst case the sliding window approach would only be expected to converge linearly, in all of our experiments (images, speech, and music), the value obtained for the optimization objective after only two passes through the signal was only a tiny fraction of a percent worse than the optimal value.[6]

### 3.2 Solving for bases

Lee et al. [13] presented an efficient way to optimize the regular sparse coding optimization problem (1-2) with respect to $a^{(j)}$. Their algorithm used the insight that the quadratic objective can be written as a sum of $p$ quadratic terms, one for each component of $a^{(j)}$. Each of these terms involves only the variables $\{a_t^{(j)}, j=1,\ldots,n\}$ for one fixed value of $t$. This therefore decomposes the optimization problem (1) into many smaller, nearly-independent problems (that are coupled only through the constraint in Equation 2). By transforming the problem into the equivalent, dual optimization problem, it could then be solved very efficiently.

Unfortunately, the same is not true for SISC. Solving the SISC optimization problem (3-4) for the bases $a$ keeping the coefficients $s$ fixed reduces to an $L_2$-constrained optimization problem:

$$\min_a \quad \sum_{i=1}^{m} \|x^{(i)} - \sum_{j=1}^{n} a^{(j)} * s^{(i,j)}\|_2^2 \quad (6)$$

$$\text{subject to} \quad \|a^{(j)}\|_2^2 \leq c, \ 1 \leq j \leq n \quad (7)$$

In this problem, because each basis function can appear in any possible shift, each component of the basis vector contributes to many different terms in the objective function. Therefore, unlike regular sparse coding, the different components of the basis functions are coupled in the objective. However, as we now demonstrate, this problem is much easier to solve when transformed into the frequency domain.

For a basis function $a = (a_1, a_2, \ldots, a_p)$, let the discrete Fourier transform (DFT) be denoted by $\hat{a} = (\hat{a}_1, \hat{a}_2, \ldots, \hat{a}_p)$, where each $\hat{a}_t$ is a complex number, representing the frequency component $t$.[7] We will use the following two facts from signal processing theory to simplify optimization problem (6-7): (i) Parseval's theorem [14] implies that the DFT $\hat{a}$ of a function $a$ scales the $L_2$ norm by a constant factor; in other words, $\|\hat{a}\|_2^2 = K\|a\|_2^2$, where $K$ is a known constant. Since our objective and constraints both consist of $L_2$ terms, we can apply the DFT to them to obtain an equivalent optimization problem over $\hat{a}$ instead of $a$. (ii) The Fourier transform of a convolution is the elementwise product of the Fourier transforms. Using (i) and (ii), and denoting the elementwise product of vectors $f$ and $g$ by $f \cdot g$, the optimization problem becomes:

$$\min_{\hat{a}} \quad \sum_i \|\hat{x}^{(i)} - \sum_j \hat{a}^{(j)} \cdot \hat{s}^{(i,j)}\|_2^2$$

$$\text{subject to} \quad \|\hat{a}^{(j)}\|_2^2 \leq \hat{c} = cK$$

where the optimization is now over the vector of complex variables $\hat{a} \in \mathbb{C}^p$, and $\hat{x}^{(i)} \in \mathbb{C}^p$ and $\hat{s}^{(i,j)} \in \mathbb{C}^p$ represent the complex-valued DFT for the input $x^{(i)}$ and coefficients $s^{(i,j)}$ respectively.[8] Now, for this problem, the Lagrangian can be decomposed as a sum of quadratic terms, each depending on a single frequency component $t$:

$$\mathcal{L}(\hat{a}, \lambda) \ = \ \sum_t \left( \|\hat{x}_t - \hat{S}_t \hat{a}_t\|_2^2 + \hat{a}_t^* \Lambda \hat{a}_t \right) - \hat{c} 1^T \lambda,$$

with dual variables $\lambda \in \mathbb{R}^n$ and

$$\hat{a}_t = \begin{pmatrix} \hat{a}_t^{(1)} \\ \vdots \\ \hat{a}_t^{(n)} \end{pmatrix}, \quad \hat{x}_t = \begin{pmatrix} \hat{x}_t^{(1)} \\ \vdots \\ \hat{x}_t^{(m)} \end{pmatrix}, \quad \Lambda = \text{diag}(\lambda),$$

$$\hat{S}_t \ = \ \begin{pmatrix} \hat{s}_t^{(1,1)} & \hat{s}_t^{(1,2)} & \cdots \\ \hat{s}_t^{(2,1)} & \hat{s}_t^{(2,2)} & \cdots \\ \vdots & \vdots & \ddots \end{pmatrix}$$

---

[6]Other details: Feature-sign search activates coefficients one at a time, because this is necessary for guaranteeing a descent direction. In our implementation, for each iteration, we simultaneously activate the $k = 300$ zero coefficients with the largest partial derivatives. This is not guaranteed to give a descent direction, but in the (extremely rare) case where it does not, we simply revert to choosing a single coefficient for that particular iteration. This preserves the theoretical convergence guarantee of feature sign.

[7]Following standard practice, we apply the DFT after padding the basis vector $a$ with zeros so as to make it the same length $p$ as the inputs $x$. In this case, the DFT is also of length $p$. Due to space constraints, we refer the reader to standard texts such as Oppenheim et al. [14] for further details on Fourier transforms.

[8]We use $\mathbb{C}$ to denote the set of complex numbers, and $y^*$ to denote the conjugate transpose of a complex vector $y \in \mathbb{C}^n$.

The Lagrangian $\mathcal{L}(\hat{a}, \lambda)$ is a function of complex variables $\hat{a}$, but can be expressed as a function of only real variables using the real and imaginary parts of $\hat{a}$. Further, since $|y|^2 = |\mathrm{R}e(y)|^2 + |\mathrm{I}m(y)|^2$ for a complex number $y$, we can analytically compute $\hat{a}^{min} = \arg\min_{\hat{a}} \mathcal{L}(\hat{a}, \hat{a})$ by optimizing over $\mathrm{R}e(\hat{a})$ and $\mathrm{I}m(\hat{a})$ to obtain:[9]

$$\hat{a}_t^{min} = \left(\hat{S}_t^* \hat{S}_t + \Lambda\right)^{-1} \hat{S}_t^* \hat{x}_t \quad (8)$$

Substituting this expression for $\hat{a}$ into the Lagrangian, we analytically derive the dual optimization problem and optimize it efficiently over the dual variables $\lambda$ using Newton's method in MATLAB's optimization toolbox.[10] Once the optimal dual variables $\lambda$ are computed, the basis can be recovered using Equation (8).

### 3.2.1 Efficient implementation

The update rules derived above rely on the matrices $\hat{S}_t$, which contain a given frequency component of the Fourier transform of the coefficient vectors $s^{(i,j)}$ for all $m$ patches and all $n$ bases. It would seem straightforward to precompute the Fourier transform of every coefficient vector and construct each $\hat{S}_t$ from its corresponding entries. Unfortunately, while the time domain coefficients $s^{(i,j)}$ are sparse, their Fourier representations are not. Therefore, as they would require $O(mn)$ storage for each frequency component, these representations cannot all be stored in memory at once for large problems. On the other hand, it is inefficient to compute each component of $\hat{S}_t$ directly from the definition of the Fourier integral each time it is used. Thus, the two most straightforward implementations of the above algorithm do not scale to large problems.

To address this, we note that the Newton updates can be computed efficiently if we have precomputed the matrices $\hat{S}_t^* \hat{S}_t$ and $\hat{S}_t^* \hat{x}_t$. Furthermore, these matrices are small enough to fit in memory, with $O(n^2)$ and $O(n)$ nonzero entries respectively for each frequency component. Importantly, unlike the $O(mn)$ required for storing the Fourier transforms of all coefficient vectors, these figures do not depend on the number of

---

[9]This can be proved by closely following the derivation for the real-valued Lagrangian used by Lee et al. [13].

[10]The update rules for Newton's method can be computed using:

$$\begin{aligned} M_t &= \hat{S}_t^* \hat{S}_t + \Lambda \\ D(\lambda) &= \sum_t \left( \|\hat{x}_t\|_2^2 - \hat{x}_t^* \hat{S}_t M_t^{-1} \hat{S}_t^* \hat{x}_t \right) - \hat{c} 1^T \lambda \\ \frac{\partial D(\lambda)}{\partial \lambda_i} &= \sum_t \|e_i^T M_t^{-1} \hat{S}_t^* \hat{x}_t\|_2^2 - \hat{c} \\ \nabla_\lambda^2 D(\lambda) &= -2 \sum_t \left( M_t^{-1} \hat{S}_t^* \hat{x}_t \hat{x}_t^* \hat{S}_t M_t^{-1} \right) \cdot \bar{M}_t^{-1}, \end{aligned}$$

where $e_i$ is the $i^{th}$ unit vector and $\cdot$ represents the Hadamard (pointwise) product.

input signals being solved for. Two further optimizations were used for our largest scale experiments: (i) Because the Fourier transform of a real signal is conjugate symmetric, $\hat{S}_t^* \hat{S}_t$ and $\hat{S}_t^* \hat{x}_t$ need to be explicitly computed and cached only for half of the frequencies $t$. (ii) Since $\hat{S}_t^* \hat{S}_t$ is Hermitian, it suffices to cache only the entries in the lower-triangular part.

## 4 Constructing features using unlabeled data

We now describe an application of SISC to self-taught learning, where the goal is to use unlabeled data to improve performance on a classification task. Here, the unlabeled data may not share the classification task's labels.

Concretely, suppose we are given a labeled training set $\{x_l^{(1)}, y_l^{(1)}, \ldots, x_l^{(k)}, y_l^{(k)}\}$ together with $m$ unlabeled examples $\{x^{(1)}, \ldots, x^{(m)}\}$. The subscript "$l$" stands for "labeled," and $y_l^{(i)}$ are the supervised learning problem's class labels. Raina et al. [6, 7] proposed the following algorithm for self-taught learning: regular sparse coding is applied to the unlabeled data $x^{(i)}$ to learn the basis functions $a^{(j)}$, by solving the optimization problem (1-2). These learnt bases are then used to construct features for each input $x_l$ to the classification task by computing:

$$f(x_l) = \arg\min_s \|x_l - \sum_j a^{(j)} s^{(j)}\|_2^2 + \beta \sum_j |s^{(j)}|.$$

When applied to images, text and music, these features $f(x_l)$ often capture slightly higher-level patterns in the input than the raw input $x_l$ itself. It is thus not surprising that when these features are used as input to standard, off-the-shelf classification algorithms such as an SVM or Gaussian discriminant analysis (GDA), they often produce better generalization performance than using the raw inputs themselves.

However, the above approach is computationally infeasible and conceptually unsatisfactory when applied to representing "large" inputs with spatial and temporal extent, such as images or audio. As a heuristic solution, in [6, 7] sparse coding was applied to small parts (or "patches") of the input images or audio signals, and then the coefficients produced for these individual parts were aggregated (e.g., by taking the sum of squares of all coefficients corresponding to a basis) to produce a representation for the entire image or audio signal. Further, the learnt feature representation $f(x_l)$ is not invariant to shifts of the input signal, and different bases might capture the same pattern occurring at different shifts.

To address these problems, we propose the following algorithm for self-taught learning using SISC: We first apply SISC to the unlabeled data $x^{(i)}$ to learn shift-invariant basis functions $a^{(j)}$. The learnt bases are then used to construct features for the labeled inputs $x_l$ by solving the SISC optimization problem:

$$\tilde{x}_l = \arg\min_s \|x_l - \sum_j a^{(j)} * s^{(j)}\|_2^2 + \beta \sum_j \|s^{(j)}\|_1.$$

Here, we will write $\tilde{x}_{l,t}^{(j)}$ to represent the coefficient found for basis $a^{(j)}$ at time $t$. As in regular sparse coding, we expect these features $\tilde{x}_l$ to capture higher-level patterns than the raw input $x_l$. Importantly, these features are also robust to shifts and translations of the inputs in the following sense: if the input signal is shifted by a certain amount in time, the features are also shifted by the same amount, *without* any change in their relative values. Using these features $\tilde{x}_l$ to represent the data, we will apply standard, off-the-shelf classification algorithms (SVM and GDA) to learn a classifier.

Concretely, GDA posits that the conditional distribution of the input given the class label is Gaussian. We use:

$$P(\tilde{x}_l|y) = \prod_t P(\tilde{x}_{l,t}|y) \quad (9)$$

where $P(\tilde{x}_{l,t}|y)$ is modeled as a multivariate Gaussian (whose parameters $\mu_y \in \mathbb{R}^n, \Sigma_y \in \mathbb{R}^{n \times n}$ depend on $y$): $[\tilde{x}_{l,t}^{(1)}, \ldots, \tilde{x}_{l,t}^{(n)}]^T \sim \mathcal{N}(\mu_y, \Sigma_y)$.

In our experiments, we found that the SISC features $\tilde{x}_l$ are poorly modeled by a multivariate Gaussian generative distribution as used in GDA—a large number of features are zero and the feature distributions have long tails.[11] Instead, the individual nonzero features appear to follow an exponential distribution. We thus propose an additional, two-step generative model for the SISC features $\tilde{x}_l$, in which first each feature's sign is determined, and if its sign is non-zero, its value is then determined according to an exponential distribution. Concretely, given $y$, for each $\tilde{x}_{l,t}^{(j)}$, we imagine that first a random variable $Z_t^{(j)} \in \{+, 0, -\}$ is drawn from a 3-way multinomial distribution governed by parameter $\phi_y^{(j)} \in \mathbb{R}^3$. $Z_t^{(j)}$ will determine the sign of $\tilde{x}_{l,t}^{(j)}$. Then, if $Z_t^{(j)} = $ "+", we generate $\tilde{x}_{l,t}^{(j)} \sim Exponential(b_{y,+}^{(j)})$; if $Z_t^{(j)} = $ "0", we set $\tilde{x}_{l,t}^{(j)} = 0$; and lastly if $Z_t^{(j)} = $ "−", we generate $-\tilde{x}_{l,t}^{(j)} \sim Exponential(b_{y,-}^{(j)})$. Here, $\phi_y^{(j)}, b_{y,+}^{(j)}$ and $b_{y,-}^{(j)}$ are the parameters of the model, and define a naive Bayes like generative model, where $P(\tilde{x}_l|y) = \prod_t \prod_j P(\tilde{x}_{l,t}^{(j)}|y)$. The maximum likelihood parameters for the generative model are straightforward to obtain,[12] and are then used to compute label probabilities $P(y|\tilde{x}_l)$ using Bayes' rule. We call this model "MultiExp," and present results on all domains using the SVM, GDA and MultiExp classification algorithms.

## 5 Experiments

### 5.1 Algorithms

We begin by evaluating our algorithm, described in Section 3.1, for solving for the coefficients given a fixed basis set. To the best of our knowledge, all previous algorithms for large-scale shift-invariant sparse coding have learned the coefficients using a heuristic to preselect a subset of coefficients to optimize for (setting all other coefficients to zero). Smith & Lewicki [11] used matching pursuit and filter threshold algorithms to select a subset of the coefficients to optimize for; we will refer to the matching pursuit heuristic as the SL heuristic. Blumensath & Davies [12] used a heuristic which iteratively chooses the coefficient with the largest magnitude gradient, and removes its "neighbors" (coefficients corresponding to the same basis with a slightly different shift) from consideration; we will call this the BD heuristic. In our implementation of each of these heuristics, we used gradient descent with line search to optimize for the selected coefficients.[13] We will compare these heuristics with our feature-sign based exact coefficient learning algorithm (FS-EXACT) and gradient descent applied to all the coefficients without any preselection (GD-FULL).[14]

Figure 2(a) shows the convergence rates of these four algorithms when applied to learning SISC coefficients for a one-second (time-domain) speech signal using 32 basis functions.[15] FS-EXACT clearly outperforms

---

[11] Some of the baseline features presented in the Experiments section, such as MFCCs, are most often modeled with a Gaussian distribution.

[12] Each multinomial parameter is estimated as a (smoothed) fraction of the occurrences of the feature with negative/zero/positive sign. Each exponential scale parameter is estimated as the absolute value of the (smoothed) average of the feature values with the corresponding negative/positive sign.

[13] Our gradient descent algorithm for solving $\min_s f(s) + \beta\|s\|_1$ (for a differentiable function $f$) transforms it to the equivalent problem $\min_{s^+,s^-} f(s^+ - s^-) + \beta 1^T(s^+ + s^-)$ with the constraints $s^+, s^- \geq 0$. The algorithm then performs a line search with exponential backoff along the negative gradient of the new objective function, handling the constraints by projecting all points considered during the line search into the feasible space. The algorithm exits when the cumulative decrease in the objective function over 3 iterations is less than a fixed threshold. This gradient descent algorithm was faster than a gradient descent algorithm using a constant step size.

[14] We do not present running time results for the sliding window algorithm FS-WINDOW as the exact algorithm FS-EXACT is usable at the scales for which results have been reported for previous algorithms.

[15] Each speech basis was 188ms long. Similar results were

the other algorithms. For example, after 8 seconds, FS-EXACT is $10^{-2}$-suboptimal, but it takes the best other algorithm 40 seconds to reach this level of suboptimality. Further, both the heuristic methods converge to suboptimal solutions.

We next assess the effectiveness of using the dual problem formulation (in the Fourier domain) for learning the bases. We compare the overall running time of basis learning using four possible algorithms, comprising combinations of two methods for solving for the coefficients—FS-EXACT and GD-FULL—and two methods for updating the bases—Newton's method with the dual formulation (DUAL) and stochastic gradient descent (GD-BASIS) on the original problem (3-4).[16] Figure 2(b) shows the convergence results for all four algorithms on the SISC image task (similar results are achieved for the speech task; details omitted due to space constraints). FS-EXACT + DUAL outperforms the other algorithms. For example, it achieves an objective value of $1.74 \times 10^5$ within 10 minutes, but the best other algorithm takes over 2 hours.

We also note that while the two basis update methods (DUAL and GD-BASIS) take comparable amounts of time per iteration, the DUAL method computes the optimal bases for the fixed coefficients at each iteration, whereas GD-BASIS just updates the bases using a small gradient step at each iteration. This contributes to faster overall convergence of basis learning using DUAL.

### 5.2 Classification

We apply our SISC based self-taught learning algorithm to two audio classification tasks: (i) Speaker identification on the TIMIT speech corpus given a small amount of labeled data (one sentence per speaker), and a large amount of unlabeled data comprising speech from *different speakers from different dialect regions* than the ones in the labeled data. (ii) Musical genre classification with a small number of labeled song snippets (6 seconds of music per genre), and a large amount of unlabeled data comprising songs from *different genres* of music than the ones in the labeled data. In both of these classification tasks, the unlabeled data is easy to obtain, and does not share the labels associated with the task. We compute SISC features for both tasks by learning bases on the unlabeled data.[17]

For speaker identification, we compare our SISC features against both spectral and cepstral features—i.e., against the raw spectrogram, and against the MFCC features that are widely used in audio applications. For music classification, we compare against using the raw spectrogram, the MFCC features, and also a set of custom, hand-engineered features designed specifically for this task by Tzanetakis & Cook [15].[18]

For each of these features, we apply the SVM, GDA and MultiExp classification algorithms to learn a classifier on the labeled training set. For an SVM, the input vector was constructed as the average of the features over the entire input signal.[19] For the generative algorithms GDA and MultiExp, each input vector was constructed by averaging the features over overlapping time "windows" of $W$ timesteps (where $W$ was picked by cross-validation). (Thus, for example, $\tilde{x}_{l,t}$ in Equation 9 is now replaced by an average of $\tilde{x}_{l,t}$ over $W$ timesteps.) Each of these windows is assumed to be independently sampled from the generative model distribution. Consequently, at test time, these algorithms output the label that maximizes the product of the generative probability for each feature window. We also augment the MultiExp algorithm by trading off the multinomial and exponential terms using an additional parameter $\alpha$, while still learning the other parameters using the usual maximum likelihood estimates.[20] The parameters of the classifiers were picked using cross-validation on a separate development set.[21]

---

obtained when a straightforward extension of our method was applied to the case of 2-D convolution for images.

[16]Note that GD-BASIS just computes a one-step update, and unlike DUAL, does not solve for the optimal basis. Using either method, one iteration of updating for the bases keeping the coefficients fixed takes only a small fraction of the time taken for learning the coefficients. Due to space constraints, we do not present results separately just for solving the bases.

[17]We picked the sparsity penalty $\beta$ for basis learning by searching for the smallest value (leading to the most nonzero coefficients) for which bases could still be learned efficiently (within 1-2 days); the same value of $\beta$ was used to construct SISC features.

[18]Tzanetakis & Cook also proposed other features based on advanced, highly specialized music concepts, such as automatic rhythm and beat detection. In this paper, we use only their 19 timbral features, since they measure "sound quality" as opposed to longer-time functions of the music signal, such as beats. The other features would still be essential for constructing a truly state-of-the-art system.

[19]For SISC features, we found it useful to count the average number of times each feature was nonzero. We also allowed SISC features to be aggregated using the average of either the square-roots, absolute values, or the squares of the individual features, picking among these using cross-validation on the development set.

[20]Using slightly informal notation, the reweighting results in writing the generative model as $P(\tilde{x}_{l,t}^{(j)}|y;\phi,b,\alpha) \propto \left(P(z_t^{(j)}|y;\phi)\right)^\alpha P(\tilde{x}_{l,t}^{(j)}|y,z_t^{(j)};b)$, where $z_t^{(j)} \in \{+,0,-\}$ corresponds to the sign of $\tilde{x}_{l,t}^{(j)}$. Such a "reweighting" term is often used in speech recognition systems to balance the language model and the acoustic model (e.g., [16]), and was used also in [17].

[21]We tuned all the standard parameters for each algorithm by cross-validation. For SVM, this includes the ker-

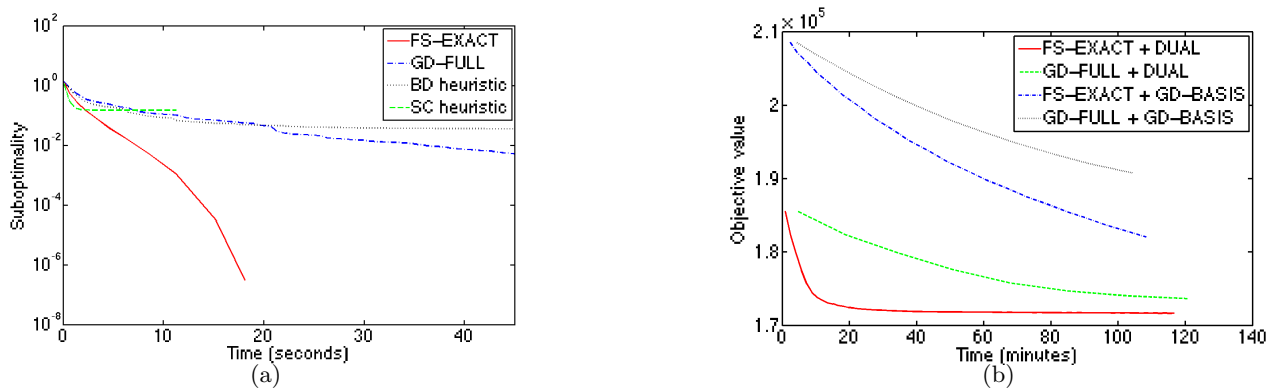

Figure 2: (a) Convergence of coefficient learning for speech data: The plot shows the "suboptimality" achieved by the algorithms vs. the time taken. (in seconds). The suboptimality of an algorithm achieving objective $f_{\text{alg}}$ for problem (5) with optimal objective $f^*$ is defined as $(f_{\text{alg}} - f^*)/f^*$. Similar results are obtained for 2-D image data. (b) Convergence of basis learning for image data: The plot shows the objective value attained by different combinations of coefficient learning and basis learning algorithms vs. the time taken (in minutes). Since the problem is not jointly convex in the coefficients and bases simultaneously, we could not compute a suboptimality relative to the global minimum. (Best viewed in color.)

| Feature Set | GDA | SVM | MultiExp |
|---:|---:|---:|---:|
| SISC | 51.0 | **56.6** | 55.7 |
| TC   | 49.3 | 47.0 | 45.8 |
| MFCC | 43.6 | 54.0 | 34.2 |
| Raw  | 44.2 | 48.4 | 42.9 |
| SISC | 37.9 | **41.8** | 39.7 |
| TC   | 38.9 | 39.2 | 37.8 |
| MFCC | 34.4 | **41.7** | 37.5 |
| Raw  | 38.8 | 38.2 | 36.3 |

Table 1: Classification accuracy on 5-way musical genre classification with 3 (top half) or 1 (bottom half) training song per genre, and three 2-second excerpts per song. For each half, the results in bold are statistically significantly better than the other results ($p \leq 0.01$).

All the reported test results are averaged over 2500 random choices of the training and test set.

Table 1 shows the test results obtained for musical genre classification with two different training set sizes.[22] The SISC features achieve the highest classification accuracy. We note that MFCCs comprise a carefully engineered feature set that has been specifically designed to capture the discriminative features of audio. In contrast, the SISC features were discovered fully *automatically* using unlabeled data, but still lead to comparable or superior performance.

Speaker identification can be done fairly accurately using standard methods when the audio samples are clean and noiseless (and when the number of speakers is relatively small), since each speaker's unique vocal tract properties often make it easy to distinguish between speakers. [16] However, the most challenging, and perhaps important, setting for speaker identification is identification even in the presence of substantial amounts of noise (for example, imagine trying to hear a person's voice on the phone over loud background noise). We are also specifically developing a speaker identification system for this setting for the STAIR (STanford AI Robot) project, as part of the robot's dialog system (see [18] for details); thus, our experiments will emphasize this more interesting, noisy setting.

We evaluated our speaker ID system under a variety of noise conditions. We recorded five kinds of household noises:[23] two electric shavers, a printer, running water, and a fan. Classification was performed under a noiseless condition, and three additional conditions: first, the same kind of noise was added to all training and test examples (SAME); second, all examples had a type of noise chosen randomly and independently (RANDOM); third, in the most difficult condition, for each speaker, all training examples contain one kind of noise, and all the test examples contain a different kind of noise (DIFFERENT).

Table 2 shows the test results for speaker identification.[24] SISC features outperform the other features

---

nel and regularization parameters; for GDA, the regularization parameter and choice of diagonal or full covariance matrix; for MultiExp, the smoothing parameters and the weight parameter $\alpha$.

[22]Other details: We applied SISC to learn 128 bases of length 500 ms from 3,000 spectrogram excerpts of length 1.5 seconds each. The spectrogram was computed with MATLAB's built-in spectrogram function, with window length 25 ms (400 samples) and 50 percent overlap; we used 300 frequencies evenly spaced from 20 to 6000 Hz. The training and test sets were always drawn from different songs.

[23]Chosen to be representative of that encounteed by STAIR. [18]

[24]Further details: We learned a set of 128 SISC bases over a log frequency spectrogram with 128 frequencies evenly spaced on a log scale from 300 to 8000 Hz. Bases were learned from dialect regions 1 through 4 of the Timit corpus, while regions 5 and 6 were used for development, and regions 7 and 8 for testing. For classification, each training and test instance consisted of a 1.5-second sample

| Noise condition | SISC | | | MFCC | | | Raw | | |
|---|---|---|---|---|---|---|---|---|---|
| | SVM | GDA | MultiExp | SVM | GDA | MultiExp | SVM | GDA | MultiExp |
| No noise | 71.2 | 74.2 | 70.7 | 46.0 | 71.6 | 62.6 | 64.9 | **78.9** | 46.1 |
| SAME, SNR=20 | 58.6 | 60.6 | 58.9 | 37.7 | 64.6 | 53.2 | 53.5 | **66.9** | 41.6 |
| SAME, SNR=10 | 55.3 | 56.3 | 53.8 | 33.5 | 57.9 | 48.6 | 50.7 | **61.8** | 40.4 |
| RANDOM, SNR=20 | 51.8 | 51.0 | **55.6** | 34.3 | **55.8** | 42.1 | 42.1 | 51.1 | 35.2 |
| RANDOM, SNR=10 | 42.0 | **48.7** | **48.7** | 29.2 | 43.8 | 31.9 | 33.1 | 38.5 | 27.5 |
| DIFFERENT, SNR=20 | 44.1 | **50.8** | **50.3** | 31.0 | 48.3 | 33.6 | 28.7 | 40.0 | 28.5 |
| DIFFERENT, SNR=10 | 32.3 | **41.0** | **41.3** | 20.3 | 31.5 | 19.7 | 21.7 | 23.3 | 27.1 |

Table 2: Classification accuracy on 5-way speaker identification, using 1 training sentence per speaker. The results in bold are statistically significantly better than the other results in the same row ($p \leq 0.01$). Higher SNR (signal/noise ratio) implies lower added noise.

under two of the three noise conditions tested.

## 6 Discussion

In this paper, we presented an efficient algorithm for SISC. Our algorithm solves for the coefficients by searching over their signs; and solves for the bases by mapping an optimization problem with highly-coupled parameters to a set of much smaller, weakly-coupled problems in the Fourier domain, thus allowing it to be solved efficiently. We also showed that this method performs well in self-taught learning, where the goal is to learn to perform a supervised learning classification task exploiting additional unlabelled data drawn from classes other than the ones of interest.

### Acknowledgments

We give warm thanks to Honglak Lee, Austin Shoemaker, Evan Smith, Stephan Stiller and the anonymous reviewers for helpful comments. This work was supported by the DARPA transfer learning program under contract number FA8750-05-2-0249.

---
from a speaker.